# Automatic individual pig detection and tracking in surveillance videos

Lei Zhang[a], Helen Gray[b], Xujiong Ye[a]*, Lisa Collins[b], Nigel Allinson[a]

[a]*Laboratory of Vision Engineering, School of Computer Science, University of Lincoln, Brayford Pool, Lincoln, LN6 7TS, UK*
[b]*Animal welefare epidemiology Lab, Faculty of Biological Sciences, University of Leeds, Leeds, LS2 9JT, UK*

**Abstract**

Individual pig detection and tracking is an important requirement in many video-based pig monitoring applications. However, it still remains a challenging task in complex scenes, due to problems of light fluctuation, similar appearances of pigs, shape deformations and occlusions. In order to tackle these problems, we propose a robust on-line multiple pig detection and tracking method which does not require manual marking or physical identification of the pigs, and works under both daylight and infrared (nighttime) light conditions. Our method couples a CNN-based detector and a correlation filter-based tracker via a novel hierarchical data association algorithm. In our method, the detector gains the best accuracy/speed trade-off by using the features derived from multiple layers at different scales in a one-stage prediction network. We define a *tag-box* for each pig as the tracking target, from which features with a more local scope are extracted for learning, and the multiple object tracking is conducted in a key-points tracking manner using learned correlation filters. Under challenging conditions, the tracking failures are modelled based on the relations between responses of detector and tracker, and the data association algorithm allows the detection hypotheses to be refined, meanwhile the drifted tracks can be corrected by probing the tracking failures followed by the re-initialization of tracking. As a result, the optimal *tracklets* can sequentially grow with on-line refined detections, and tracking fragments are correctly integrated into respective tracks while keeping the original identifications. Experiments with a dataset captured from a commercial farm show that our method can robustly detect and track multiple pigs under challenging conditions. The promising performance of the proposed method also demonstrates a feasibility of long-term individual pig tracking in a complex environment and thus promises a commercial potential.

## 1. Introduction

Pork is the world's most consumed meat product, with approximately 770 million pigs farmed globally each year, producing over 110 million tons of meat (USDA, 2017). Monitoring pig health is crucial for both productivity and welfare, but can be challenging given the large scale of many farms. Current surveillance methods of pig disease and

Click here to enter text.

behavior in the UK often involve human observation, either as daily checks by farm staff or more in depth quarterly veterinary assessments. This practice is not only time consuming, but may result in inaccurate observations, given that animals are known to change or cease the behaviour of interest in the presence of a human observer [1-3]. It is almost impossible to observe animals on a continuous and individual basis using just human observers, and this is even more the case on a commercial farm. Moreover, each separate observer may have a bias in how they interpret behaviours or clinical signs, leading to a potentially unreliable representation of individual and pen-level activity [4]. In addition, many declines in health begin with a change in activity level and/or a decrease in food and water consumption (e.g. endemic diarrhoea, bursitis, lameness and porcine reproductive and respiratory syndrome [PRRS]) [5]; signs which are difficult to spot in a brief 'snapshot' visit to a pen. In some cases, it is challenging to manually identify illnesses which can take effect very quickly. For example, conditions of the nervous system, such as meningitis or bowel oedema, may develop from an unnoticed case to the onset of body convulsions and death within four hours [5]. The swift development of such conditions means that the critical time point for disease identification may fall between inspection times. The ability to automatically detect and track the movement of individual pigs over an extended period could therefore aid in the early detection of potential health or welfare problems without the need for human observation.

Several studies have already begun to use automatic surveillance techniques to observe pig behaviour. For example, there has been success with the use of RFID ear tags to monitor pig movement [6]. However, due to the large number of (passive or active) RFIDs readers and tags needed, the utility of RFIDs, especially active RFIDs, are compromised as an expensive surveillance option. The attachment and detachment of tags also entails additional labour costs. For these reasons, tagging is not routinely applied on commercial farms. A more promising solution for pig monitoring is using camera-based computer vision techniques, as such approaches provide a low cost, non-invasive and non-attached solution. Three dimensional video cameras (top view with depth sensors) have been used to monitor pigs in several studies [7-10]. The point clouds from depth sensors are processed and analysed to detect and track pigs. However, the accuracy of pig detection and tracking using depth sensors is sensitive to the quality of the depth images and given performance limitations of existing depth sensors (e.g. Kinect), good depth images can only be captured over a limited range and field of view, this in turn could complicate the sensor installation on a large-scale farm. The more conventional 2D gray/colour video camera is the most commonly used solution in many video surveillance systems. Over the past twenty years, a few 2D video camera-based pig segmentation/detection and tracking methods have been developed. For example some state-of-the-art image processing techniques such as GMM-based background subtraction [11, 12], denoising using low-pass filtering followed by Otsu's thresholding, morphological operations and ellipse fitting [13-15], graphical module based segmentation [16], and learning-based tracking [17] have been applied. However, most of these 3D/2D camera-based vision approaches for pig monitoring have not been discussed in relation to problematic scenarios and practical difficulties in a harsh environment on a real-world commercial farm. There are three major problems with video-based detection and tracking of individual pigs (see Fig. 1): 1) light fluctuation. e.g. sudden light changes frequently occur in the pig sheds, including different illuminations during day and night (light-on and light off) that can trigger a change in the camera imaging model (e.g. from Fig. 1. a under the normal model to Fig. 1. d under IR model). This creates unavoidable shadowing in the different light conditions (Fig. 1 d). 2) Very similar appearances of pigs and varying status of background. 3) object deformations and occlusions. e.g. insect transitorily occludes the lens (Fig. 1.c, f), pigs crowd together and occlude each other (Fig. 1.a, b, c, e). In this last case, despite individual pig detection and tracking being considered an essential stage in many video-based pig monitoring applications, developing a system that can cope with these condition remains an open challenge.

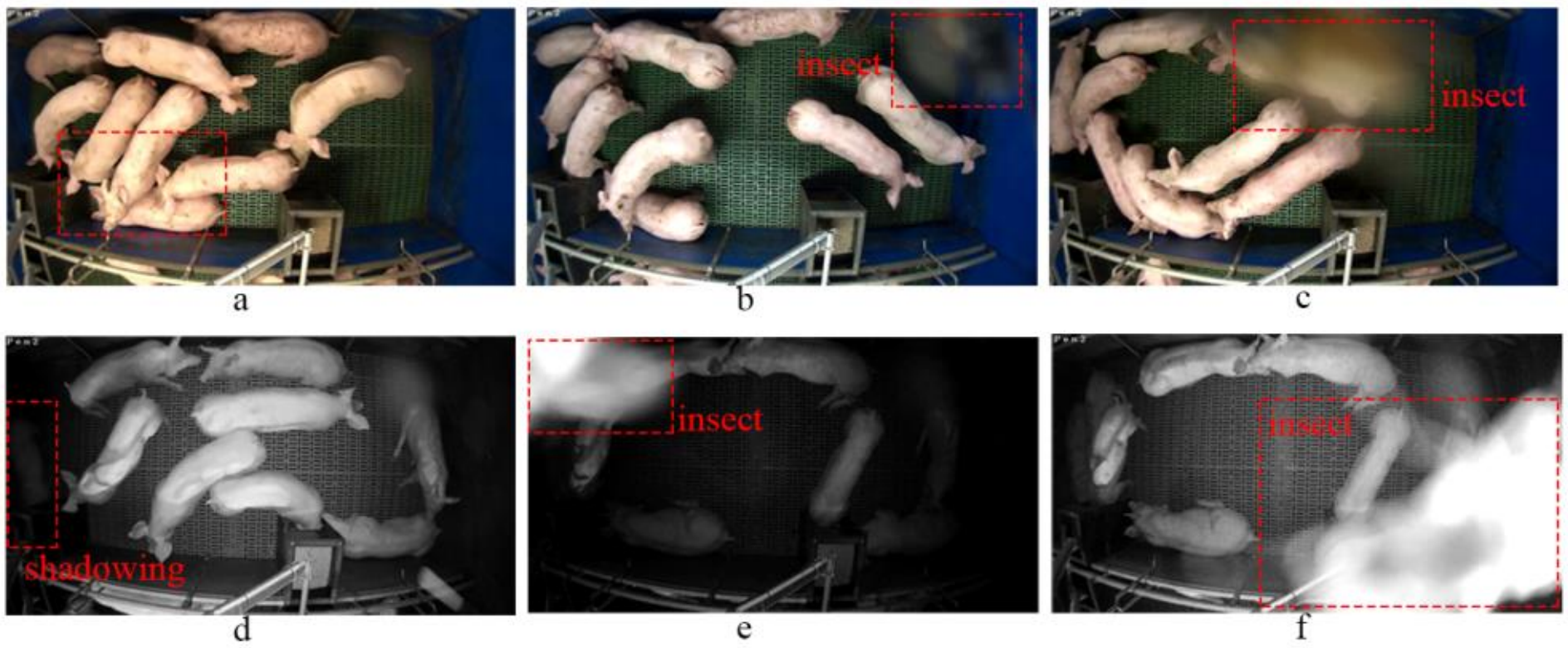

Fig. 1. Challenges of automatic detection and tracking for individual pigs in real-world scenarios on a commercial farm. (a) Occlusions by pigs in day time (light-on) video footages; (b) insect causes the changes of the illuminations from frame a to b in day time footages; (c) Occlusions by insect in light-on footages; (d) shadows in the light-off video footages under IR model. (e) the insect effects on light-off footages; (f) the occlusion by insect in light-off footages.

To tackle these problems, we present a 2D video camera-based method to automatically detect and track individual pigs in an indoor shed, without the need to manually mark or physically identify pigs. The system works on both colour video and grayscale video under daylight and IR light conditions, respectively, which also has the capability of tackling occlusion problems caused by insects and pigs. Unlike previous attempts to detect pigs (group level or individual level) using traditional image processing algorithms (e.g. GMM background subtraction and thresholding method), the hierarchical features derived from a deep learning architecture are used to detect each pig in the pen. We focus on investigating the features of the foreground objects (pigs) rather than the background, given the varying status of the background in different farming systems (e.g. slatted floors and straw-based systems). Given the visual similarities of pig appearances in the captured footage from top-view cameras, it is difficult to extract an intrinsic feature to identify each pig (e.g. using face recognition techniques). In our method, individual pig detection is defined as a multiple object detection problem while fully considering complicated scenes. The fundamental idea in multiple pig tracking is to continuously update a model for each pig using discriminative correlation filters [18]. In order to tackle the occlusion problem that could result in incorrect detections and so make the tracking fail, we propose a data association algorithm to complementarily bridge the detection and tracking processes based on the contextual cues in sequential frames. To the best of our knowledge, no attempt has been made to detect and track individual pigs in such complicated scenes and it is the first attempt of using deep learning methods for pig detection and tracking at an individual level.

The remainder of this paper is organised as follows. Section 2 reviews previous automatic vision-based pig detection and tracking methods. The experimental materials and our method are presented in Section 3. Section 4 reports on our experimental results. Further discussion is presented in Section 5. We conclude the paper in the final section.

## 2. Related work

In the past twenty years, several methods using 2D video cameras have been introduced for detecting and tracking pigs. For example, McFarlane et al. [13] proposed a piglet segmentation and tracking algorithm. The segmentation method was a combination of image differencing with respect to a median background and a Laplacian operator, and during the tracking, each piglet was modelled by an ellipse which is calculated from the blob edges in the segmented image. The major problem in segmentation is the limitation of distinguishing tightly grouped piglets, which leads to tracking errors. Moreover, large, sudden movements of the piglets could cause tracking failure. Kashiha et al. [14] proposed an automated method to identify marked pigs in a pen using pattern recognition techniques. The pigs were firstly segmented from the image by denoising the image using a 2D

Gaussian filter followed by Otsu's thresholding. Then the marks on the pigs were extracted within a similar segmentation manner, which was further used to identify the pattern based on a Fourier Description. This method requires the pigs to be individually marked, which is not always practical on a commercial scale. Moreover, the major problem of this method is that paint patterns will fade out over time, meanwhile the movements of pig could result in unclear paint patterns, and in some conditions (e.g. light-off or night environment), the marks become invisible. A study in the literature [15] investigated the change in group lying behaviour of pigs related to changing environmental temperature. The Otsu's segmentation, morphological operations and ellipse fitting were used to segment the pigs from the background, and the Delaunay triangulation method employed to analyse the group behaviour of pigs. Nilsson et al. [19] proposed a learning-based pig segmentation method to measure the proportion of pigs located in different areas. Several state-of-the-art features were extracted from the image to feed into a logistic regression solver using elastic net regularization. A typical failure case using this method is due to the occlusions in the pen (e.g. occlusions by the wall and pigs occluding each other to some degree). Ahrendt et al. [17] developed an individual pig tracking algorithm in loose-housed stables. The tracking algorithm consists of two stages: preliminary pig segments were firstly generated using background subtraction followed by thresholding, then the spatial and RGB information derived from the pig segments in a frame were used to build up a 5-dimensional Gaussian model for each individual pig. The major limitation of this method is that the tracking could be lost if the movements of the pigs were rapid, and also when occlusion occurs among pigs. Tu et al. [16] proposed a pig detection algorithm in grey-scale video footage, where the pigs were detected by foreground object segmentation. The method consists of three stages, in the first stage the texture information is integrated to update background modelling, in the second stage, pseudo-wavelet coefficients are calculated which are further used in the final stage to approximate a probability map using a factor graph with a second-order neighbourhood system and loopy belief propagation (BP) algorithm. The disadvantage of this method is the considerable high computational complexity due to using factor graphs and a BP algorithm. In the literature [11], the Gaussian Mixture model (GMM) based background subtraction method [20] is used to detect moving pigs under windowless and 24-hour light-on conditions. However, it is augured that the GMM background is time-consuming. In order to overcome this problem of traditional GMM, Li et.al [12] proposed an improved algorithm based on an adaptive Gaussian mixture model for pig detection, where the Gaussian distribution was scanned once every *m* frames and the excessive Gaussian distribution is detected to improve the convergence speed of the model. The major problem of this method is that the detection could fail when sudden lighting changes occurred.

Recently, 3D video cameras (top-view based-depth sensor) have been used to monitor pigs in several studies [7-10]. For example, Kulikov et al. [7] proposed 3D imaging sensor-based method to automatically track piglets in a field test, in which several Microsoft Kinect 3D image sensors and 3D image reconstruction using EthoStudio software were adopted. Another proposed system [8] assessed the normal walking patterns of pigs, tracking the trajectories using a 6-camera Vicon system with reflective markers [21] and a Microsoft Kinect camera. Kim et al. [9] proposed a method to detect standing pigs using a Kinect camera, in which the temporal noise in the depth images is removed by applying spatiotemporal interpolation, then detection is achieved by applying edge detection based on the undefined depth values around pigs. Matthews et al. [10] proposed an individual pig detection and tracking method based on the depth maps generated from a Kinect camera, whereby a region-growing approach is applied on the pre-processed images (with calibration, noise removal and surface normalization) to detect the pigs and tracking is implemented by linking detection in adjacent frames using the Hungarian Assignment algorithm. Although existing depth video camera-based vision systems have achieved some successes in pig detection and tracking, they tend to have certain drawbacks due to the limitation of the imaging sensor (e.g. Kinect depth camera has limited range (4 m) and field of view (horizontal 58.5° and vertical 45.6°)), and the accuracy of depth data is sensitive to camera position. This, in turn, complicates device installation for large-scale commercial agricultural application. Furthermore, due to the variety of the shed or pen sizes in different commercial farms, this physical limitation of the depth sensor could significantly reduce the system generalization. In addition, regardless of noise in the point clouds, a 3D top-view sensor cannot address the occlusion problem.

In this study, we focus only on 2D camera-based applications related to detection and tracking of individual pigs.

## 3. The proposed method

We first present a brief overview of our proposed method, Fig. 2 shows a block diagram of the method, which consists of three main components: object detection, multiple object tracking and data association.

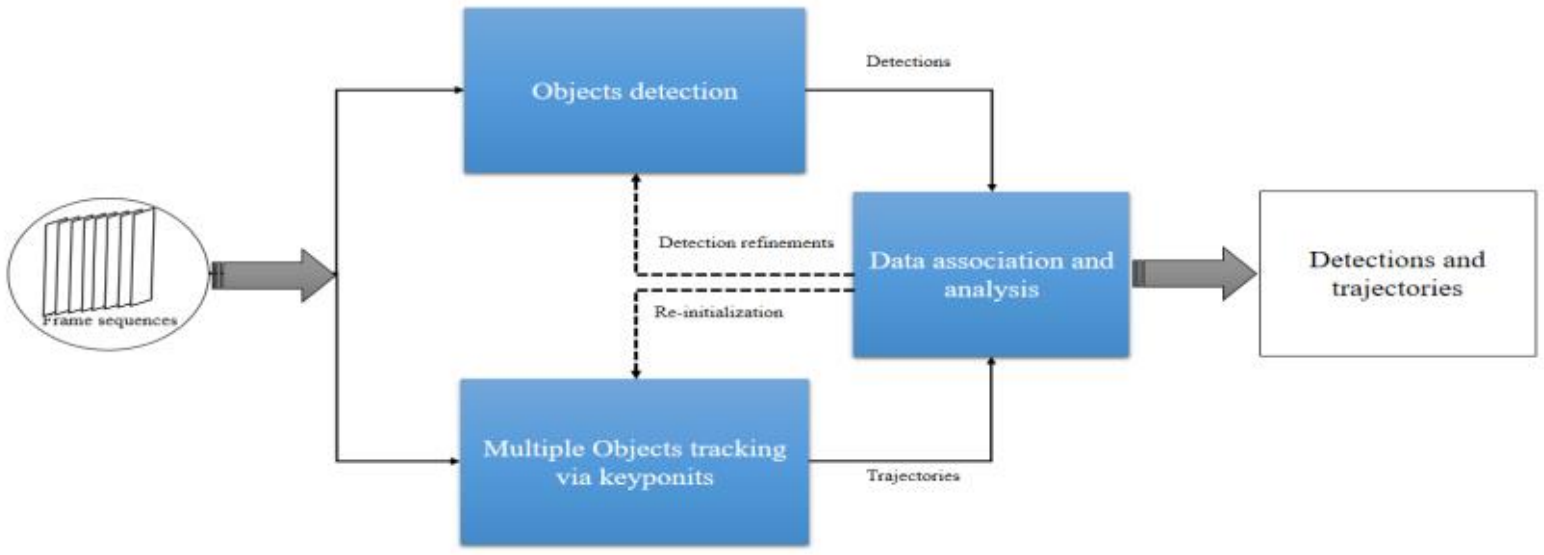

Fig. 2. The general framework of proposed method

In the object detection, the detector works independently on each individual frame to localize the areas that appear similar to the training samples. In many object detection pipelines, the key task is to extract representative features from the foreground (objects), which tends to make it more reliable in dealing with illumination variation and occlusion compared to background subtraction. In the multiple object tracking process, each object in the scene is assigned a unique ID, the tracker predicts motions of all object instances between consecutive frames and uses the same unique ID for each object throughout its appearance in the scene. Given the complicated pen environment in commercial farms, both detector and tracker may fail in a video sequence. For instance, the false positives and false negatives may appear in detections while the trajectories of object instances can be fragmented or interrupted during the tracking. The data association process is used to refine the detections based on temporal context and to form robust trajectories for each object instance under the assumption that the object motions between the adjacent frames are limited. It also allows the detection and tracking to work in a complementary manner. When a large tracking error is detected (e.g. the objects are invisible due to an occlusion), the system will automatically trigger the re-initialization of tracking obtained from the detector. By this data association, both detector and tracker are coupled to predict accurate object bounding boxes with more coherent trajectories.

### *3.1. CNN based Object detection*

Object detection is a crucial middle level task in computer vision, which facilitates high level tasks such as action recognition and behavioural analysis. It is also a prerequisite stage to 'tracking by detection' based object tracking methods. Since convolutional neural networks (CNN) were introduced into the object detection community, due to their relatively superior performances compared to the traditional hand-crafted features-based detectors, CNN-based object detection has received considerable attention, and a great deal of progress has been made in recent years. Current state-of-the-art CNN based object detection methods [22-25] mostly follow the pioneering work of the R-CNN [26]. The CNN network in the R-CNN can be considered as a feature extractor, working on the regions which are cropped from an input image by some external region process (e.g. selective search [27]). Classification and localization are solved by a support vector machine (SVM) classifier and a class-specific bonding box (BB) regression. However, R-CNNs are slow because each region proposal needs to go through a CNN forward pass, where overlapping Regions of Interest (RoIs) proposals lead to significant duplicated computation. The problem is mitigated in the Fast R-CNN architecture [22], which takes the entire image once through the network to obtain a feature map. Then, for each region proposal, a RoI pooling layer extracts fixed resolution regional features from the feature map. A multi-task loss, $L$, which is a combination of classification and regression loss is used to train each labeled RoI:

$$L(p,c,t^c,g) = \alpha \cdot l_{class}(p,c) + \beta \cdot f(c) \cdot l_{reg}(t^c,g) \quad (1)$$

where, each training RoI is labeled with a truth class $c \in 0, 1, 2 \ldots K$ over $K$+1 classes, and c=0 indicates background, and a ground-truth bounding box regression target $g$ associated with the class $c$. $p = (p_0, \ldots, p_k)$ is the predicted probability distribution for each RoI computed by the soft-max. $t^c$ denotes predicted bounding box regression offsets for class $c$ over the $K$ classes, which is represented by the four parameterised coordinates of the predicted bounding box. The parameterization of these four coordinates follows the work of [26]. The *f(c)* is an Iverson bracket indicator function [c≥1], $f(x)$=1 if $c \geq 1$ and $f(x)$=0 otherwise. $L_{class}(p,c)$ and $l_{reg}(t^c,t)$ are the classification loss and regression loss, respectively, where the $L_{class}(p, c) = -\log p_c$ *is* log loss for ground truth class c, and $l_{reg}(t^c,t)$ is expressed by

$$l_{reg}(t^c,g) = \sum_{i \in \{x,y,w,h\}} smooth_{L_1}(t_i^c - g_i) \quad (2)$$

In which, the $smooth_{L_1}(x)$ is the robust smooth L1 loss function defined as $smooth_{L_1}(x) = 0.5x^2$ if $|x| < 1$, otherwise $smooth_{L_1}(x) = |x| - 0.5$. The classification and bounding box regression are balanced via the parameter $\alpha$ and $\beta$.

Both R-CNN and Fast R-CNN require an external initial RoI proposal paradigm that prevents the network being trained in an end-to-end manner. Moreover, these CNN-based detection methods, which rely on a specified external RoI, are not efficient enough to satisfy many vision applications. The follow-up work on Faster R-CNN [24] is much more efficient than the Fast R-CNN, this is mainly from the Regional proposal network (RPN) which generates initial RoI regions for subsequent learning. The Faster R-CNN has two main stages to train: in the first stage, the features extracted by a feed-forward CNN at an intermediate level are used to generate RoI proposals. The RoI proposals are generated using a sliding window on the feature map at different scales and aspect radios that act as ‘anchors’. The term ‘anchors’ is also functionally identical to the ‘default box’ in other related work [25]. In the second stage, the box proposals along with the intermediate feature map generated in the RoI proposal stage are fed to train a weight sum of classification loss and bounding box regression loss. The loss functions in both stages take the form of equation 1.

Although the Faster R-CNN has shown remarkable accuracy of detection, it is still far from being a real-time system. This motivated improvements in the detection speed while maintaining the detection accuracy in follow up work on Region-based Fully Convolutional Networks (R-FCN) [23] and Single Shot Multibox Detector (SSD) [25]. The R-FCN has a similar architecture compared to the Faster R-CNN, which consists of region proposal stage, class prediction and bounding box regression stage. It improves the efficiency of detection by reducing the computations in each region. This is achieved by extracting per-region features in the feature map generated from the deeper (upper) layer which is prior to the predictions, rather than on the feature map derived from the RPN. The SSD is the fastest architecture among these state-of-art architectures while it has competitive detection accuracy compared to the Faster R-CNN and R-FCN. The SSD uses a single feed-forward CNN to directly predict a set of anchors (default) boxes and the confidence of presence of an object class in those boxes. The improvement in speed results from using the RPN like network only for predictions without requiring an additional second stage per-region pixel or feature resampling operations. In our experiments, we implemented and compared the Faster R-CNN, R-FCN and SSD for pig detection, and adopted the SSD architecture [25] with a modified loss function as the detector (Fig. 2). Figure 3 shows the overall architecture of the SSD network.

The overall network consists of two parts: the front truncated backbone network (VGG-16) which is an image classification network (including layers: Conv1_1, Conv1_2…., Conv4_3…Conv5_3) and the additional convolutional feature layers (Conv8_1, Conv8_2… Conv11_1, Conv11_2) which progressively decrease in size. It is different from the RPN in that multiple feature maps from upper (including added feature layers Conv7, Conv8_2, Conv9_2, Conv10_2, Conv10_2 in Fig. 3) and lower layers (Conv4_3) form a single network used for default box generation as well as confidence and location predictions. More specifically, a set of default boxes (anchors) are tiled with a convolution manner at different scales and aspect ratios for these feature maps. Given the number of $f$ feature maps, the scale of the default boxes for each feature map is computed as:

$$s_i = s_{\min} + \frac{s_{\max} - s_{\min}}{f-1}(i-1), i \in [1, f] \quad (3)$$

where $S_{min}$ and $S_{max}$ denote the scales in the lowest layer and highest layer, and the scales of all the other layers in between are increased by a constant factor. A set of default boxes with $k$ different aspect ratios (e.g. 1, 2, 3, 1/2, 1/3) are associated with each grid on each feature map. For example, given a feature map with size of $m \times n$, in which each grid is located $k$=6 *or* $k$=4 default boxes, there are $m \times n \times k$ default boxes for this feature map. Therefore, there are a total of 8,732 default boxes for all feature maps (see Fig. 3). Meanwhile, two separate sets of small convolutional filters (3x3) are applied to each feature map to produce confidence and location predictions (bounding boxes offsets related to default boxes), respectively. The outputs (location predictions, confidence predictions and default box coordinates) generated from different feature layers are then concatenated separately, and subsequently fed to the Multi-task Loss unit.

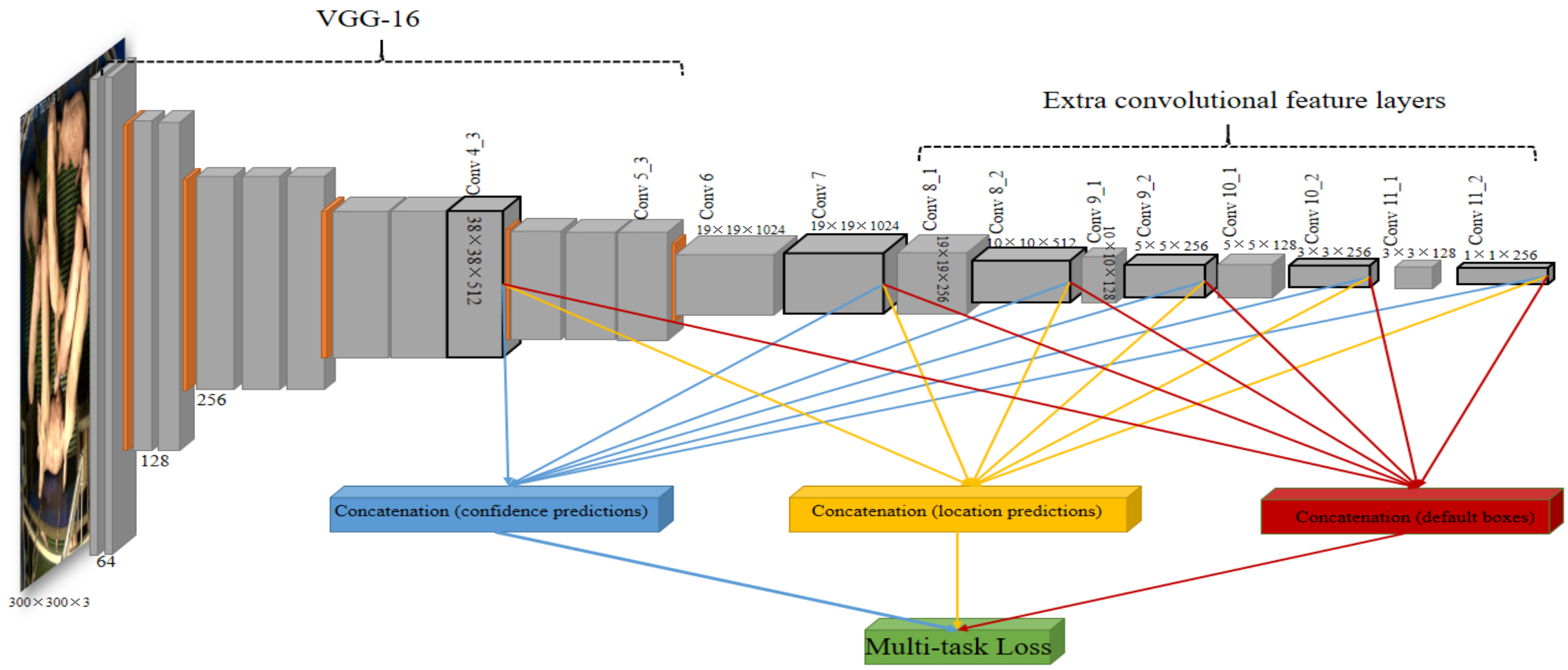

Fig. 3. The architecture of SSD object detection

During training, a two-stage matching strategy is first applied to generate positive and negative default boxes for subsequent learning. Matching a GT box to the closest default box is based on the maximum Jaccard distance overlap followed by matching default boxes to the GT box with Jaccard distance overlap higher than a threshold of 0.5. If such a match can be found, it is considered as positive default box, otherwise, it is termed a negative default box. Therefore, matching in this many-to-one manner, each GT box could have multiple positive default box samples. Meanwhile, hard negative mining is used to balance the positive and negative default boxes to the ratio of 1:3. This should prevent an imbalanced classes learning problem, and makes the network more convergent. The model is trained by minimizing an objective function, which is a mixed classification and regression loss function. In our experiment, given the scenario that the desired detection objects (pigs) are pre-determined, only two classes (pig-object and background) need to be predicted. Therefore, our loss function for an image is defined as:

$$L(p_i, c_i^*, t_i, g_i) = \frac{1}{N} \cdot \left( \sum_i^N l_{class}(p_i, c_i^*) + \beta \cdot \sum_i^N c_i^* \cdot l_{reg}(t_i, g_i) \right) \quad (4)$$

where $N$ is the number of matched default boxes; $p_i$ denotes the probability of $i^{th}$ default box in an image predicted as an object. The $c_i^*$ is the corresponding ground-truth label: if the default box is positive $c_i^*$ =1, otherwise $c_i^*$ =0 if the default box is negative. $t_i$ and $g_i$ are the predicted box parameters and the ground truth box parameters. Instead of using softmaxloss as confidence loss function for multiple classes classification, we define the $l_{class}$ as log loss over two classes (pig and background). For the regression loss, we adopted the same smooth $L_1$ loss function defined in Eqn. (2).

### *3.2. Multiple objects tracking using correlation filter*

Multiple object tracking (MOT) is the task of discovering multiple objects and estimating the trajectory of each target in a scene. In general, a MOT method can be categorized as on-line or offline tracking. The difference between these two is that the on-line tracking requires only the information from past frames when inferring states of objects in the current frame while the offline tracking requires additional information in the future frames for the inference in the current frame. Given the on-line nature of the surveillance system, the on-line tracking manner is appropriate for our task. In this paper, given sequential frames for individual pig tracking, let $s_t^i$ be the state of the $i_{th}$ pig in the $t_{th}$ frame, and let $S_t = \left(s_t^1, s_t^2 ..., s_t^N\right)$ be all states of $N$ pigs in the $t_{th}$ frame. For each pig in the video, the sequential states of the $i_{th}$ pig are denoted by $s_{1:t}^i = \left\{s_1^i, s_2^i, ..., s_t^i\right\}$. Thus, $S_{1:t} = \left\{S_1, S_2 ..., S_t\right\}$ denotes all the sequential states of all pigs in a video. The objective of the individual pig tracking is to find the optimal sequential states of all $N$ pigs by various learning-based methods based on the existing training samples. In our method, the Discriminative Correlation Filters (DCF) based *on-line* tracking method [28] is adopted to track each pig.

The DCF has shown its advances in object tracking in recent years. Since the MOSSE[29] was first introduced for adaptive tracking using DCF, a series of improvements [28, 30, 31] have been made to this tracker to achieve more accurate, robust and efficient tracking. These methods learn a continuous multiple channel convolution filter from a set of training samples which are collected from the past (e.g. $1{:}(t\text{-}1)_{th}$) frames. The object is tracked by correlating the learned filter over a search window in the current ($t_{th}$) frame, where the location has maximal responses to the filter indicates the new position. Benefited from periodic assumption of the training samples the training and prediction are efficient in the frequency domain by applying the Fast Fourier Transform (FFT). Formally, given $M$ training samples $\left\{x_j\right\}_1^M$, each sample $x_j$ consists of $D$ feature maps ($x^1_j$, $x^2_j$,…, $x^D_j$) extracted from an image region using one or more object appearance modeling methods (such as HOG, Colour names and deep CNN features etc.), of which each $x^d_j$ has an independent resolution. A set of convolution filters $f = (f^1, f^2, …, f^D)$ is trained to predict the confidence scores $R_f\left\{x\right\}$ of the target, and the target is localized by maximizing the $R_f\left\{x\right\}$

$$R_f\left\{x\right\} = \sum_{d=1}^{D} f^d * H_d\left\{x^d\right\} \tag{5}$$

where $*$ indicates the convolution operator. In order to fuse the multiple-resolution feature maps in the tracking, the feature map is transferred to the continuous spatial domain by an implicit interpolation model given by operator $H_d$. In the training phase, given a set of $M$ sample pairs $\left\{\left(x_j, \; y_j\right)\right\}_1^M$, the filters $f$ are learned by minimizing the following objective,

$$\underset{f}{\arg\min} \sum_{j=1}^{M} \alpha_j \left\| R_f\left\{x_j\right\} - y_j \right\|^2 + \sum_{d=1}^{D} \left\| w \cdot f^d \right\| \tag{6}$$

in which, the samples $x_j$ are labeled by desired confidence scores $y_j$, defined in the continuous spatial domain, as the periodic repetition of the Gaussian function. $\alpha_j$ indicates the weight for each sample $x_j$, and the classification error between the $R_f\{x_j\}$ of sample $x_j$ and corresponding labels is given by the L2 norm. The additional term in Equation (6) is a spatial regularization component to suppress unwanted boundary effects, where $w$ is a regularization matrix containing weights determining the importance of the filter coefficients. To train the filters, Eqn. (6) is minimized in the Fourier domain and the filters are iteratively updated by employing a Conjugate gradient (CG) method [31].

In order to speed up individual pig tracking and tackle the occlusion problems in the scene, we follow the improvements [28] based on the aforementioned tracker with three aspects: reducing the number of weights in the trainable filters, reducing the sampling redundancy, and adopting a sparser model updating scheme. The number of weights is reduced by introducing a factored convolution operator given by:

$$R_{Pf}\{x\} = \sum_{c,d} p_{d,c} f^c * H_d\{x^d\} = f * P^T H\{x\} \tag{7}$$

The $f^d$ in *D-dimensional* filters $f = (f^1, f^2, \ldots, f^D)$ in Eqn. (5) are constructed as a linear combination of a smaller set of basis filters $f^c$ *in* $(f^1, f^2, \ldots, f^C)$ using a set of learned coefficients which is represented as a $D \times C$ matrix $P = p_{d,c}$. Then the feature vector $H\{x\}$ at each location $t$ is multiplied with the matrix $P^T$ followed by convolving with the filters to calculate the confidence scores. This results in compacting the $D$-dimensional filter to $C$-dimensional filter thus that reduces the model size and improves the efficiency of tracking. During the training, the filter $f$ and matrix $P$ are learned jointly via Gauss-Newton and the conjugate gradient method.

In the sampling stage, instead of constructing a sample set containing $M$ samples by collecting a new sample in each frame, the sample set is constructed based on the joint probability distribution $p(x,y)$ of the sample feature maps $x$ and corresponding desired confidence scores $y$. Assuming the target is centered in its search window of the image, all $y$ scores can be predefined as the same Gaussian function $y^*$, thus, $p(x,y)$ can be approximated by a Gaussian Mixture model (GMM) $p(x)$. The GMM is constructed by a weighted sum of $L$ Gaussian components which is given by:

$$P(x) = \sum_{l=1}^{L} \pi_l g(x, \mu_{l,} I) \tag{8}$$

where $\mu_{l,}$ is the desired Gaussian mean to each component $l$, and $\pi_l$ is its weight determined by the learning rate, and the covariance matrix $I$ is set to the identity matrix. The GMM model is updated by adopting the simplified on-line learning algorithm [32]. This GMM sample modeling scheme results in more efficient learning by reducing the number of samples from $M$ to $L$. Meanwhile it is the variety of the samples that enhances the accuracy of representing the samples. This is a particularly important factor that allows the method to cope with problems related to the deformations and occlusions of pigs. By using GMM, then the objective function in (6) is transformed to the following objective:

$$\underset{f}{\arg\min} \sum_{l=1}^{L} \pi_l \left\| R_f\{\mu_l\} - y^* \right\|^2 + \sum_{d=1}^{D} \left\| w \cdot f^d \right\| \tag{9}$$

In order to further improve the tracking capability of handling the occlusions (e.g. insects) and deformations, the same sparse filters updating scheme [28] is adopted in the learning. Namely, rather than updating the filters in each frame, the filters are updated on a fixed interval of frames and the loss is updated by adding a new component that encodes the features from a set of object varieties. The infrequent updating helps to reduce the over-fitting as well as improving learning stability.

In our scenario, the target is relatively large (see fig.4) that leads to high computation cost of extracting features and convoluting filters with such large sample. As a result, the efficiency of the tracking will be significantly reduced. Meanwhile, given the similarity of appearances among pigs and the severe overlapping in a pen, the feature with a wider scope (e.g. the entire pig) is not discriminative enough that could increase tracking drifts under occlusions. Therefore, in our method, we define the target as a portion of the pig body (e.g. the center green box in fig. 4 is named as *tag-box*) and the aim is to track this *tag-box* by features extracted in local scope. These targets (*tag-box*) are tractable based on the key points tracking manner. This is benefited from that the aforementioned tracking method with continuous formulation (equation 5) enables accurate sub-pixel localization.

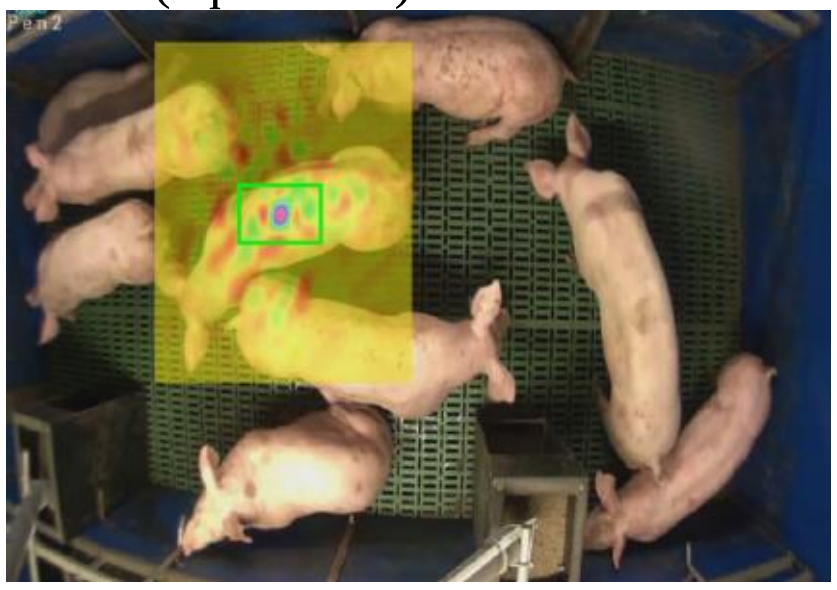

Fig. 4. The tracking target (green box) in our method and its search area (yellow box)

### *3.3. Data association*

In our method, we assume that both the detector and tracker described in the above sections may fail in some environments due to occlusions and deformations. For example, the false positives and false negatives may occur in the detections and tracking drift, or fragmentation of trajectories may be caused by the occlusions. In order to improve the robustness of the detection and tracking, especially for long-term tracking of multiple pigs under occlusions, we propose a novel hierarchical data association algorithm to bridge the detector and tracker. The association is formulated as a bipartite graph matching and solved using the Hungarian algorithm [33]. Thus, the MOT problem in our method is treated as a pairwise association by associating detector responses with tracklets, which can recover tracking failures according to the estimated object hypothesis from the detections. In order to recover tracking failures, we need to determine in which condition the object is observed. In our method, the different conditions are expressed by the relation between the detection bounding box and the *tag-box* of the tracker described above. More specifically, let $DB_t = \{Db_t^1, Db_t^2, Db_t^3 ... Db_t^J\}$ and $TB_t = \{Tb_t^1, Tb_t^2, Tb_t^3 ... Tb_t^N\}$ which are a set of detection bounding boxes and a set of *tag-boxes* respectively in the $t_{\mathrm{th}}$ frame , where $Db_t^j$ denotes the $j_{\mathrm{th}}$ detection bounding box in the $t_{\mathrm{th}}$ frame and $Tb_t^i$ is the $i_{\mathrm{th}}$ *tag-box* in the $t_{\mathrm{th}}$ frame. Four states are defined according to the $DB_t$ and $TB_t$ , for the object under different conditions that could lead to tracking failures, the corresponding responses are triggered and implemented to recover the tracking.

(1). An object is tracked if a $Db_t^j$ contains only one $Tb_t^i$ , which means the one-to-one association between the detection bounding boxes and the *tag-boxes* has been established. (State: tracked)

(2). An object is not currently tracked due to an occlusion if the $Tb_t^i$ is not assigned to any existing tracks, namely the *tag-box* of tracker is located out of the default detection bounding box $Db_{t-l}^{\bullet i}$ . This will trigger a *tag-box* initialization scheme. (State: tracking drift [tracking target shifts away from the detection bounding box])

(3). Unstable detection occurs if the $Db_{t-l}^{\bullet i}$ which is restrained by the $Tb_t^i$ is not assigned to any $Db_t^j$ . This will trigger a detection refinement scheme based on the tracklet derived from the historical detections. (State: unstable detection).

(4). If more than one $Tb_t^i$ are assigned to a $Db_t^j$ , it means the tag-box with less assignment confidence is drifting to an associated bounding box. This triggers a *tag-box* pending process followed by the initialization in the condition (2). (State: tracking drift)

Given the above four states, in general, the MOT with occlusion problems can be simplified to assess the relations between the detection bounding boxes and tracking tag-boxes. Detector and tracker work in a complementary manner, namely, detection can be refined by the historical tracks which are restrained by the tracked *tag-box* while the tracking failures can be successfully recovered by reinitializing the *tag-box* based on the optimal detections. As a result, the optimal tracklets can sequentially grow from a set of stable detections under occlusion conditions. Algorithm 1 gives a brief outline of our data association method, where step 1, 15, and 24 in the procedure are related to the pairwise associations.

**Algorithm 1** our hierarchical data association algorithm

Input: Current $t_{\text{th}}$ frame and previous trajectories
The detection bounding boxes computed from the detector described in the section 3.1
The *tag-boxes* computed from the tracker described in the section 3.2

Output: bounding boxes and trajectories for the $t_{\text{th}}$ frame

**Procedure**

**1** associate the detection bounding boxes $DB_t$ to the *tag-boxes* $TB_t$
2 **If** each $Db_t^j$ in $DB_t$ is assigned to the corresponding $Tb_t^i$ in $TB_t$ with one-to-one manner
3 Return the $DB_t$ as tracked bounding boxes $DB_t^{\bullet}$ and update the tracklets
4 **else**
5 Update the tracklets related to the $DB_t^{\bullet}$ and return the assigned detection boxes ( $DB_t^{\bullet}$ ), unassigned detection boxes ( $DB_t^{*}$ ) and unassigned *tag-boxes* ( $TB_t^{*}$ ).
6 **end if**
7 **for** each *tag-box* $Tb_t^i$ in the unassigned *tag-boxes* $TB_t^{*}$
8 Set a default box ( $Db_{t-l}^{\bullet i}$ ) to the *tag-box* $Tb_t^i$ according to the updated tracklets
**9** associate the unsigned detection boxes ( $DB_t^{*}$ ) to the default box ( $Db_{t-l}^{\bullet i}$ )
10 **if** find a best matched box in the $DB_t^{*}$
11 Set the best matched box as the tracked bounding box and update the tracklet
12 **else**
13 Set the default box as the tracked bounding box and update the tracklet
14 **end if**
**15** associate the unassigned *tag-boxes* ( $TB_t^{*}$ ) to the default box ( $Db_{t-l}^{\bullet i}$ )
16 **if** no matched *tag-box* is founded
17 Set a counter array (*age*), *age*[ $Tb_t^i$ ]=*age*[ $Tb_t^i$ ]+1
18 **else**
19 *age*[ $Tb_t^i$ ]=*age*[ $Tb_t^i$ ]-1
20 **end if**
21 **if** *age*[ $Tb_t^i$ ]>threshold value (*T*)
22 Initialize the *tag-box* $Tb_t^i$ and reset *age*[ $Tb_t^i$ ]=0;
23 **end if**
**24** associate the assigned detection boxes ( $DB_t^{\bullet}$ ) to the unassigned *tag-box* $Tb_t^i$
25 **if** the assigned detection box has more than one *tag-boxes*
26 pend the *tag-box* $Tb_t^i$
27 **end if**
28 **end for**

The association is performed by constructing an association cost matrix between detection bounding boxes and tracking *tag-box*es followed by an optimisation process using the Hungarian algorithm to determine optimal matching pairs. The cost function for a pair in the cost matrix is defined as:

$$C(Db_t^j, Tb_t^i) = \begin{cases} -\log(overlap(Db_t^j, Tb_t^i)) + \delta \cdot \dfrac{d(Db_t^j(c_{x,y}), Tb_t^i(c_{x,y}))}{diag(Db_t^j)} & if\ \ 0 < overlap(Db_t^j, Tb_t^i) \le 1 \\ 1 & if\ \ overlap(Db_t^j, Tb_t^i) = 0 \end{cases} \tag{10}$$

$$overlap(Db_t^j, Tb_t^i) = \frac{Db_t^j \cap Tb_t^i}{Tb_t^i} \tag{11}$$

where the first *overlap* term in (10) evaluates the relation between detection and *tag-box*, and the second term integrates the relative spatial information into the formula. The weight $\delta$ restrains the impact of the second term to the cost. The function *d*(.) is Euclidean distance between two boxes' center points $Db_t^j(c_{x,y}), Tb_t^i(c_{x,y})$ . and $diag(Db_t^j)$ is the diagonal length of the bounding box $Db_t^j$ . The lower cost of the function is the more likely that the affinity of two boxes is high.

## 4. Experiments and results

### *4.1. Materials and evaluation metrics*

The dataset used was collected from Spen Farm, University of Leeds, UK. Nine finisher pigs (Large White x Landrace breed) were filmed in pens of size 400 × 176 cm, with fully-slatted floors. The lighting in the shed is manually operated, with artificial light between the hours of 07:30 -16:00 and with additional natural light from two small windows. The footage was continually recorded over three days, covering day and night (examples are shown in Fig.1). The frame resolution is 1920 × 1080 pixels and the video captured with a frame rate of 20 frames/s. We screened the footage to obtain the most representative scenarios for different conditions described above (e.g. light fluctuation, object deformations and occlusions). In our experiments, there are a total of 18,000 frames for algorithm training and a total of 4,200 frames coming from five different sequences for testing. The training and testing samples contained examples of all challenging scenarios. Table 1 summarizes the details of the testing dataset.

Table 1. A summary of the testing sequences.

| Sequence | Model | Number of frames | The conditions of the sequence |
|---|---|---|---|
| S1 | day | 800 | Deformations, and light fluctuation, occlusions caused by pigs and a long stay insect on the camera |
| S2 | day | 700 | Deformations, severe occlusions happened among pigs that results in the object instances becoming invisible during the occlusions |
| S3 | day | 1500 | Deformations, and light fluctuation, occlusions caused by an insect, occlusions among the pigs |
| S4 | Night | 600 | Deformations, occlusions caused by an insect, occlusions among the pigs |
| S5 | Night | 600 | Deformations, occlusions among the pigs |

An expert provided the ground truth by manually annotating location and identification for each pig in each frame of all training and testing sequences. These annotations were used to train the algorithm and validate the proposed method.

Since the problem of individual pig detection and tracking is a typical MOT problem, we employed standard MOT evaluation metrics [34] with additional metrics (MT PT ML) used in [35] to validate our method. This set of metrics is described in the Table 2. Here, ↑ indicates that the performance is better if the number is higher, and ↓ denotes that lower number indicates a better performance.

Table 2. The evaluation metrics to assess our method

| Metric | Description |
|---|---|
| Recall ↑ | Percentage of correctly matched detections to ground-truth detections |
| Precision↑ | Percentage of correctly matched detections to total detections |
| FAF↓ | Number of false alarms per frame |
| MT↑, PT, ML↓ | Number of mostly tracked, partially tracked and mostly lost trajectories |
| IDs↓ | Number of identity switches |
| FRA↓ | Number of the fragmentations of trajectories |
| MOTA↑ | The overall multiple objects tracking accuracy |

### *4.2. Implementation details*

We implemented the individual pig detection and tracking method using Matlab and the MatConvNet CNN library [36] on a PC with configuration: Inter(R) Core(TM) i9-7900x CPU@3.30 GHZ, 32 GB RAM and Geforce GTX 1080Ti GPU. The videos were captured by a Longse-400 IP IR Dome camera powered by POE with a fixed 2.8 mm lens, the codec used was H265 (also known as HEVC).

For the system described in Section 3.1, three CNN detection architectures (Faster-RCNN, R-FCN and SSD) were implemented and compared. The backbone architecture for the three detection networks is VGG16. In the Faster-RCNN and R-FCN, feature maps used to predict region proposals are derived from the conv5_3 in the VGG16 as shown in the Fig.3. During the training, in the Faster-RCNN and R-FCN, the networks are trained on input images scaled to M pixels on the shorter edge. In SSD, the input images are resized into a fixed shape with M×M. In our experiment, we set the M=300. The networks are trained using SGD with a 0.001 initial learning rate, 0.9 momentum and 0.0005 weight decay. The batch size was chosen as 16. For the correlation based tracker, we adopted the HoG [37] and Colour Names (CN) [38] feature representation to model the appearances for the daytime model sequences, and only HoG is used in the nighttime model sequences due to the nature of the gray scale frames. In the experiment, the cell sizes of the HoG and CN were 6 and 4, respectively. For the factor correlation filter, we used the same number of CG iterations for the filters updating after the first frame. For the sampling strategy, we set the learning rate to 0.09 and the number of components is set to 30. The model is updated every two frames. In the first frame of each sequence, the initial detection bounding boxes are manually delineated and the tag-boxes are initialized to be automatically centered at the bounding boxes. We use ten Gauss-Newton iterations and 120 CG iterations to optimize the filter coefficients in the first frame. In our experiments, all parameter settings were kept fixed for all sequences in the dataset. In order to optimize the efficiency of our method, the computation of the detections was allocated to the GPU while the tracking is performed on the CPU in parallel computing manner.

### *4.3. Experimental results*

We compared three detection architectures as described in section 3.1. Fig.5 shows detection results of these three networks. It can be observed that the SSD has a relatively better detection performance compared to the R-FCN and Faster R-CNN. In Fig. 5(a), nine pigs are detected from the scene with minor occlusions. However, the detected bounding box for the pig at the periphery of the pen is narrowed. Although this pig is well localized in the Faster-RCNN in the Fig.5 (b), two overlapped pigs on the left side of the pen are improperly detected to one pig. This results in only eight pigs are detected by the Faster R-CNN. The SSD produces the best detection bounding boxes with high probability scores shown in the Fig.5 (c). This benefits from the SSD generating much more default boxes compared to the Faster R-CNN and R-FCN. Moreover, the dual matching strategy in SSD allows the network to predict high scores for multiple overlapping default boxes. The speeds of the detection in our experiments for the R-FCN, Faster R-CNN and SSD are 98 ms, 130 ms and 420 ms per frame, respectively. Given that the SSD provides the best trade-off between accuracy and speed, it is particularly suited to our application.

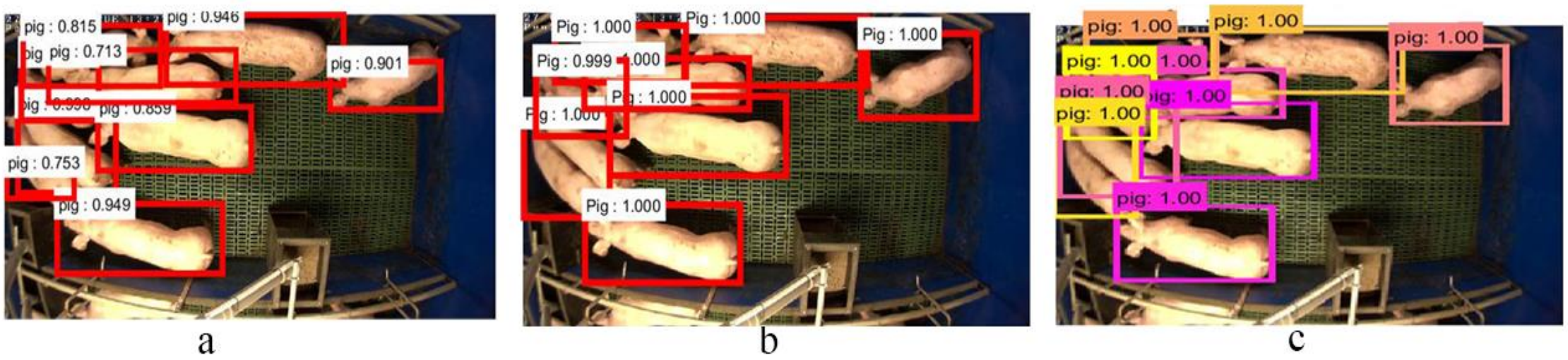

a b c

Fig. 5. Comparisons of different detection architectures. The detections results shown in (a) (b) (c) are produced from the R-FCN, the Faster-RCNN and the SSD, respectively.

As we mentioned above that detection could be fail, as the detector treats each frame as independent. More importantly, the shape deformations and light fluctuations have a significant influence that leads to the false positives and false negatives in the detections. Our method shows a robust detection and tracking performance to tackle these issues. Fig.6 illustrates an example detection and tracking result of our method. We can observe the light changes and the shape variations of the target (ID 3 pig) in the sequence from frame *a* to *g*. Even with the pig shape deforming and light changing over this time, ID 3 is successfully detected and tracked in each frame. The scale of the *tag-box* is adaptive to the changes of the target over the sequence. It is notable that our method has robust performance even in a scenario shown in the Fig.6 (f), where the shape of the pig is severely distorted and two pigs are overlapped.

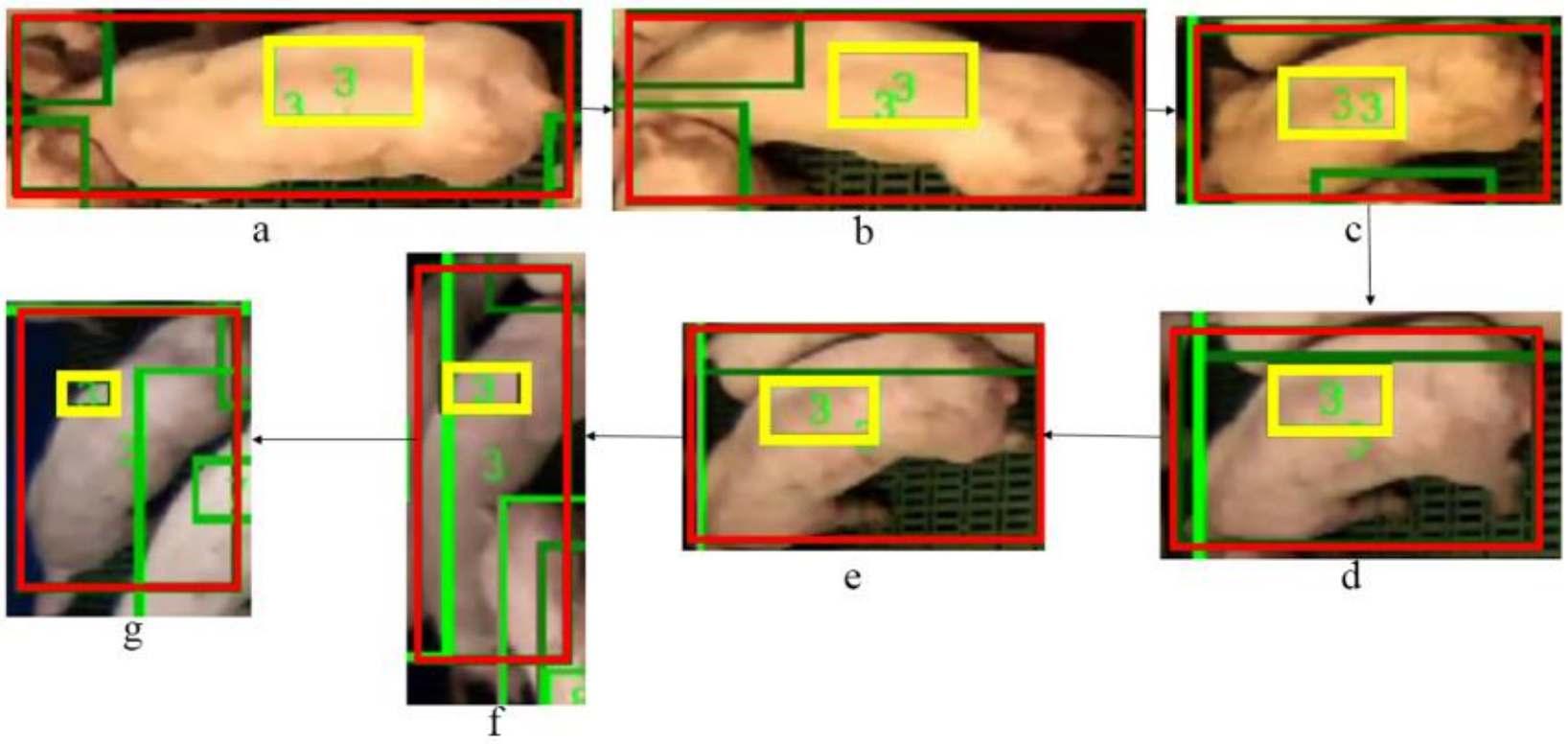

Fig. 6. The example results of our method to tackle the shape deformations. The final detected and tracked bounding boxes are shown in red colour, and the tag-boxes are shown in yellow. The sequence order is from frame a to g. ID 3 is correctly kept with correct detection bounding boxes.

The drift of the *tag-box* tracking is mainly caused by occlusions. Our method also possesses a robustness in detection and tracking under occlusion condition. For example, Fig. 7 (from frame *a* to *h*) shows an example of detection and tracking results for the occlusions with an insect. As we can see from the Fig.7 (b), an insect breaks into the scene that results in the occlusion of pig ID 2. The insect continues further across the scene in the frame Fig. 7 (c), worsening the occlusions. As a result, the occlusion directly leads to the number 6 *tag-box* drift. We can see the *tag-box* of ID 6 adheres to the local area around the bottom of the pig ID 2 in the subsequent frames (Fig.7 d, e, f). By triggering the re-initialization of the drifted ID 6 *tag-box*, it has successfully recovered in the Fig.7 (g), and the system continues tracking in the subsequent frame after re-initialization, see the Fig.7 (h).

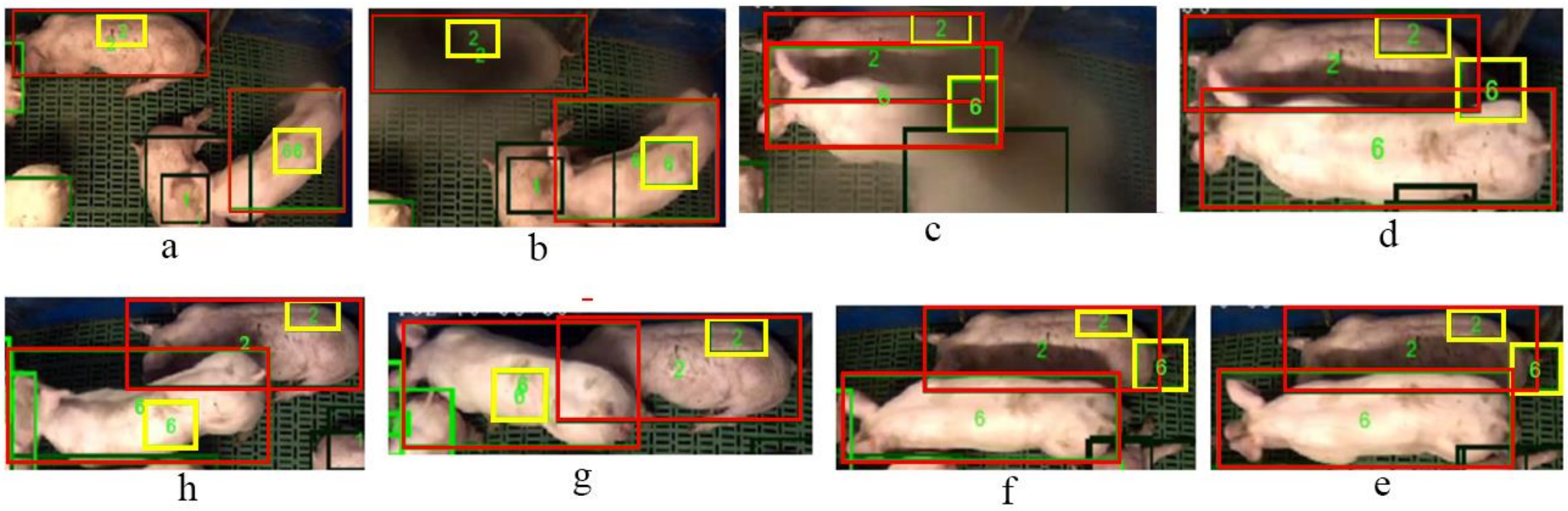

Fig. 7. The example results of our method to tackle the occlusion of insect. The final detected and tracked bounding boxes are shown in red colour, and the tag-boxes are shown in yellow. The sequences order is from frame a to h.

The main advantage of our method is that it can handle severe occlusions for the detection and tracking of multiple pigs within both nighttime and daytime conditions (Figs. 8 and 9). For example, for a night video, we can see the pigs are detected and tracked correctly in the Fig.8 frame *a* to *c*. In the frames from *d* to *f*, an insect crosses over the camera that results in a sudden light change as well as the low contrast in the frames (comparing frame *b* to *g*). More specifically, in the frame *d*, as the bounding box with ID 6 contains more than one tag-boxes (ID 6 and ID 5) while the drifted tag-box of the ID 5 (illustrated by yellow box in the Fig.8 d) is not correctly assigned to the corresponding detection box. This condition triggers the pending scheme that the tag-box of ID 5 is pended in the frames from *e* to *g*, the pending process further triggers *tag-box* initialization process in the frame *h*. In addition, we can see from the Fig.8 that the pig of ID 1 lies on the left side of the pen under the shadow in the frame *a*, and the condition becomes more complicated when this pig moves close to the pig of ID 1 and lies next to it (see fig.8 d, e, f, g). Even in such challenging condition, our method still produces stable detection and tracking results for the pig of ID 1 in every frame of the sequence. The detection and tracking results in the Fig. 9 also shows the robustness of our method to tackle the severe occlusion of pigs in day videos.

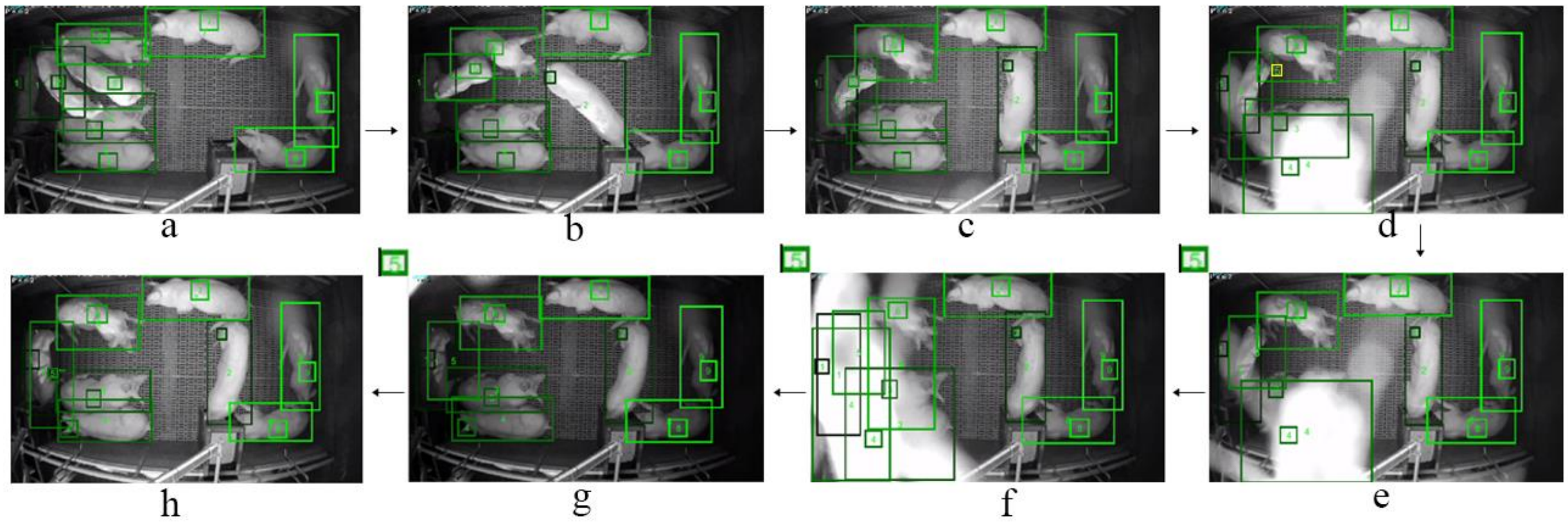

Fig. 8. The example detection and tracking results of our method in night video with a severe occlusion of insect. The sequence order is from frame a to h.

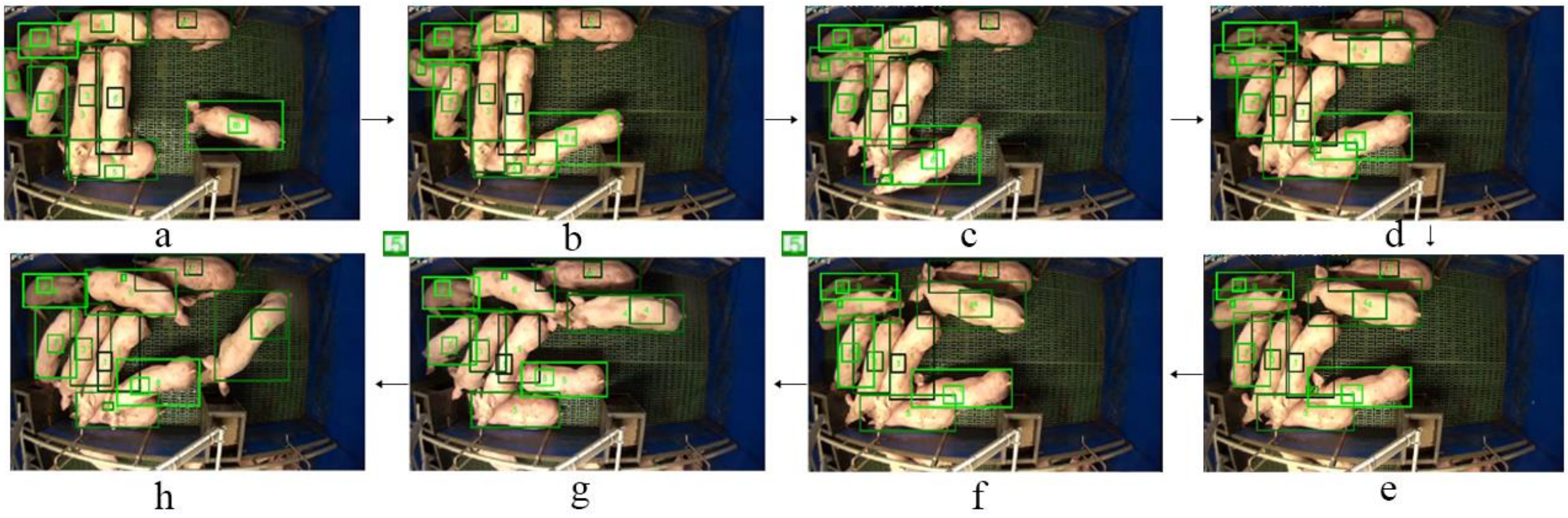

Fig. 9. The example detection and tracking results of our method in day video with a severe occlusion related to the pigs overlapping. The sequence order is from frame a to h.

Fig. 10 illustrates trajectories of nine pigs which are predicted from the S3 test sequence. The trajectories are plotted using the central points of the bounding boxes in the sequence. They are distinguished by different colors associated with different pig IDs. With some interferences (e.g. shape deformations, light fluctuation and insect), a number of track fragments can be observed. However, our method correctly associated these track fragments into nine integrated trajectories.

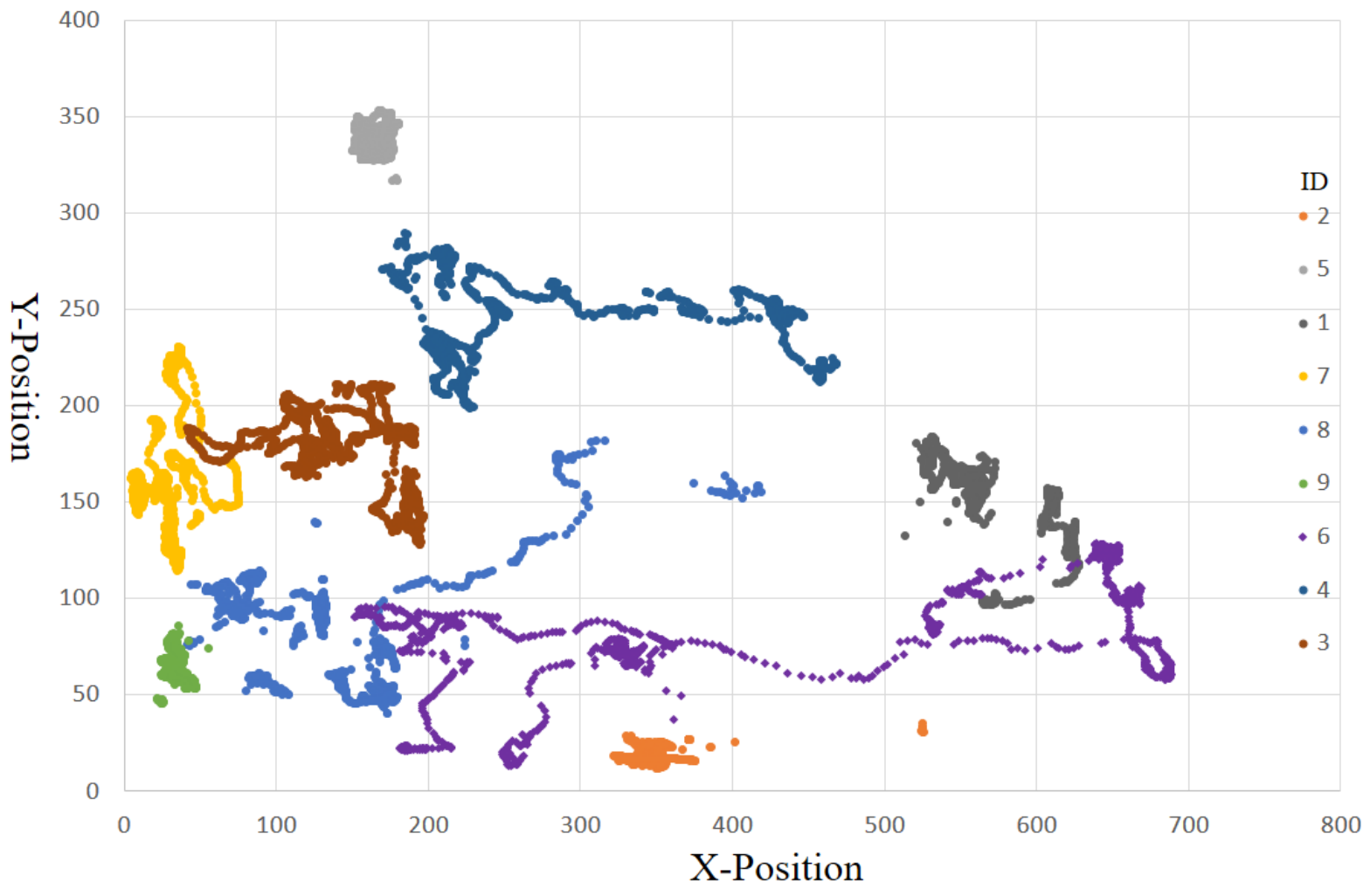

Fig. 10. The predicted trajectories of nine pigs using our method.

We evaluated our method using MOT evaluation metrics described in the section 4.1. The quantitative analysis of our method is presented in the table 3. As our method employs the tracking-by-detection strategy, both detection and tracking performances are measured in our experiments. Overall, our method achieves precision of 94.72%, recall of 94.74%, and MOTA of 89.58%. This implies that our method can robustly detect and track multiple pigs under challenging conditions. The relatively low FAF of 0.47 indicates that our method can reduce the false positives in detections to tackle the three major challenges. It is notable that the metric of the mostly lost targets (ML) is zero. This implies all nine pigs are tracked appropriately. Only one pig in the sequence S1 is partially tracked. The relatively low number of ID switches and FRA of 66.2 indicates our method can robustly construct tracks under challenging conditions. The MT of 8.8 with ML of 0 and IDs of 18 also demonstrate that the track fragments are well integrated into respective tracks while their original IDs are recovered correctly in the scenes under the occlusions, shape deformations and light fluctuations conditions.

Table 3. Evaluation results of our method for individual pigs detections and tracking

| Sequences ID | Recall (%)↑ | Precision (%)↑ | FAF ↓ | MT↑ | PT | ML ↓ | IDs ↓ | FRA ↓ | MOTA(%)↑ |
|---|---|---|---|---|---|---|---|---|---|
| S1 | 91.51 | 91.95 | 0.72 | 8 | 1 | 0 | 28 | 106 | 83.9 |
| S2 | 94.12 | 93.98 | 0.54 | 9 | 0 | 0 | 30 | 123 | 88.1 |
| S3 | 97.64 | 97.57 | 0.22 | 9 | 0 | 0 | 10 | 52 | **95.2** |
| S4 | 92.90 | 92.79 | 0.65 | 9 | 0 | 0 | 14 | 26 | 85.7 |
| S5 | 97.52 | 97.30 | **0.21** | 9 | 0 | 0 | **8** | **24** | 95.0 |
| Average | 94.74 | 94.72 | 0.47 | 8.8 | 0.2 | 0 | 18 | 66.2 | 89.58 |

## 5. Discussion and further work

With regard to the detectors described in section 3.1, it could be argued that the performances of the R-FCN and Faster R-CNN can be improved by adjusting more proposed regions in the RPN. However, due to the two-stage prediction in these networks, the operational speed depends on the number of proposals. Increasing the number of the boxes can significantly reduce the speed. Speed can be increased by using fewer boxes, but at the risk of reducing accuracy. It can be difficult to decide which detector is best suited to a particular application. In this case, we investigated the three most representative CNN based detection architectures, and found that the SSD displays the best accuracy/speed trade-off for multiple pig detections.

The similar appearances of pigs can critically affect the tacking. To some extent, reliable tracking performance using a correlation filter-based tracker relies on the discrimination in the appearance model. Although there are many features proposed for object tracking, we find pixel-wise features with a more local scope are more suitable than region features which are extracted from a wider range to track multiple similar objects. Moreover, due to the similarity of the colour as well as the shapes between pigs, the features extracted based on colour and shape cues may not be discriminative enough to construct reliable tracks. In most cases, conducting these less discriminative features for multiple object tracking can result in serious track drift. Therefore, in our method, we applied the HoG features on the more local region (tag-box) instead of on the wider region (bounding box). Of course, it is possible to track the objects with bounding boxes by increasing the number of cells in the HoG; however, this larger region can significantly increase the computational cost and so reduce the tracking speed. It may be worthwhile to investigate more hand-crafted features extracted from more local regions (e.g. Haar, Surf, ORB, etc.).

We proposed a data association algorithm to bridge the detector and tracker that allows them to work in a complementary manner. The detection and tracking confidence (failures) for a target is represented by the spatial relations between the *tag-box* and the detection bounding box for the first time. Using the data association, the detector and tracker are coupled to predict more accurate object detection with more coherent trajectories. It is notable that the spatial relation defined in the formula (10) is a straightforward way to model the tracking confidence, however, more sophisticated and efficient motion features can be integrated into the pipeline by defining a different cost function based on additional cues (e.g. velocity and acceleration).

## 6. Conclusion

In this paper, we present an automatic multiple object detection and tracking method to monitor individual pigs in a challenging environment. The proposed method performs individual pig detection and tracking without the need to manually mark or physically identify the pigs, working under both daylight and nighttime conditions. In order to tackle major problems (e.g., light fluctuation, similar appearances, object deformations and occlusions), which can significantly lead to detection failures and increasing track fragments and track drifts, we propose a method which couples a CNN based detector and a correlation filter-based tracker via a hierarchical data association algorithm. For the detections, by using the hierarchical features derived from multiple layers at different scales in a one-stage prediction network, we obtain the best accuracy/speed trade-off. For the correlation filter-based tracker, the target is defined as the tag-box which is a portion of the pig, and the tag-boxes are tracked based on the learned correlation filters working on key points tracking. Under severe conditions, the data association algorithm allows the detection hypotheses to be refined; meanwhile the drifted tracks can be corrected by probing the tracking failures followed by the re-initialization of tracking. The tracking failures are modified by the relationship between the responses of the detector and tracker. In conclusion, the data association algorithm, the detector and tracker are coupled to predict more accurate object instance inference with more coherent trajectories. Our method is trained and evaluated on 22,200 frames captured from a commercial farm. Overall, the evaluation results in a precision of 94.72%, recall of 94.74%, and MOTA of 89.58% show that our method can robustly detect and track multiple pigs under challenging conditions.

## Acknowledgements

The work is supported by the grant “PigSustain: assessing the resilience of the UK pig industry” (BB/N020790/1), which is funded by the Global Food Security ' Food System Resilience’ Programme, which is supported by BBSRC, NERC, ESRC and the Scottish Government. The authors would like to acknowledge the help of staff at the University of Leeds’ Spen Farm.